\def\year{2019}\relax
\documentclass[letterpaper]{article} %
\usepackage{aaai19}  %
\usepackage{times}  %
\usepackage{helvet}  %
\usepackage{courier}  %
\usepackage{url}  %
\usepackage{graphicx}  %
\usepackage{tablefootnote}
\usepackage{verbatim}
\usepackage{booktabs}
\usepackage{setspace}
\usepackage{xspace}
\usepackage{natbib}
\usepackage{multirow}
\usepackage{todonotes}
\newcommand{\figref}[1]{Fig.~\ref{#1}}
\newcommand{\tabref}[1]{Tab.~\ref{#1}}

\newcommand{\fst}{1\textsuperscript{st}}
\newcommand{\snd}{2\textsuperscript{nd}}
\newcommand{\trd}{3\textsuperscript{rd}}
\newcommand{\parabank}{\textsc{ParaBank}\xspace}
\newcommand{\paranmt}{\textsc{ParaNMT}\xspace}
\newcommand{\redbold}[1]{\textcolor{red}{\textbf{#1}}}
\newcommand{\bluebold}[1]{\textcolor{blue}{\textbf{#1}}}

\begin{document}
\title{\parabank: Monolingual Bitext Generation and Sentential Paraphrasing\\via Lexically-constrained Neural Machine Translation}
\author{J. Edward Hu \hspace{.25cm} Rachel Rudinger 
\hspace{.25cm} Matt Post \hspace{.25cm} Benjamin Van Durme\\
3400 North Charles Street\\
Johns Hopkins University\\
Baltimore, MD, USA{}
}
\maketitle
\begin{abstract}
We present \parabank, a large-scale English paraphrase dataset that surpasses prior work in both quantity and quality.
Following the approach of \paranmt \citep{paranmt}, we train a Czech-English neural machine translation (NMT) system to generate novel paraphrases of English reference sentences.
By adding lexical constraints to the NMT decoding procedure, however, we are able to produce \textit{multiple} high-quality sentential paraphrases per source sentence, yielding an English paraphrase resource with more than 4 billion generated tokens and exhibiting greater lexical diversity.
Using human judgments, we also demonstrate that \parabank's paraphrases improve over \paranmt on both semantic similarity and fluency.
Finally, we use \parabank to train a monolingual NMT model with the same support for lexically-constrained decoding for sentence rewriting tasks.
\end{abstract}

\setcounter{secnumdepth}{2}

\section{Introduction}

In natural languages, mappings between meaning and
utterance may be many-to-many.
Just as ambiguity allows for multiple semantic interpretations of a single sentence, a single meaning can be realized by
different sentences.
The ability to identify and generate  \emph{paraphrases} has been pursued in the context of many natural language processing (NLP) tasks, e.g., semantic similarity, plagiarism detection, translation evaluation, monolingual transduction tasks such as text simplification and style transfer, textual entailment, and short-answer grading.

\begin{figure}
  \centering
    \includegraphics[width=0.47\textwidth]{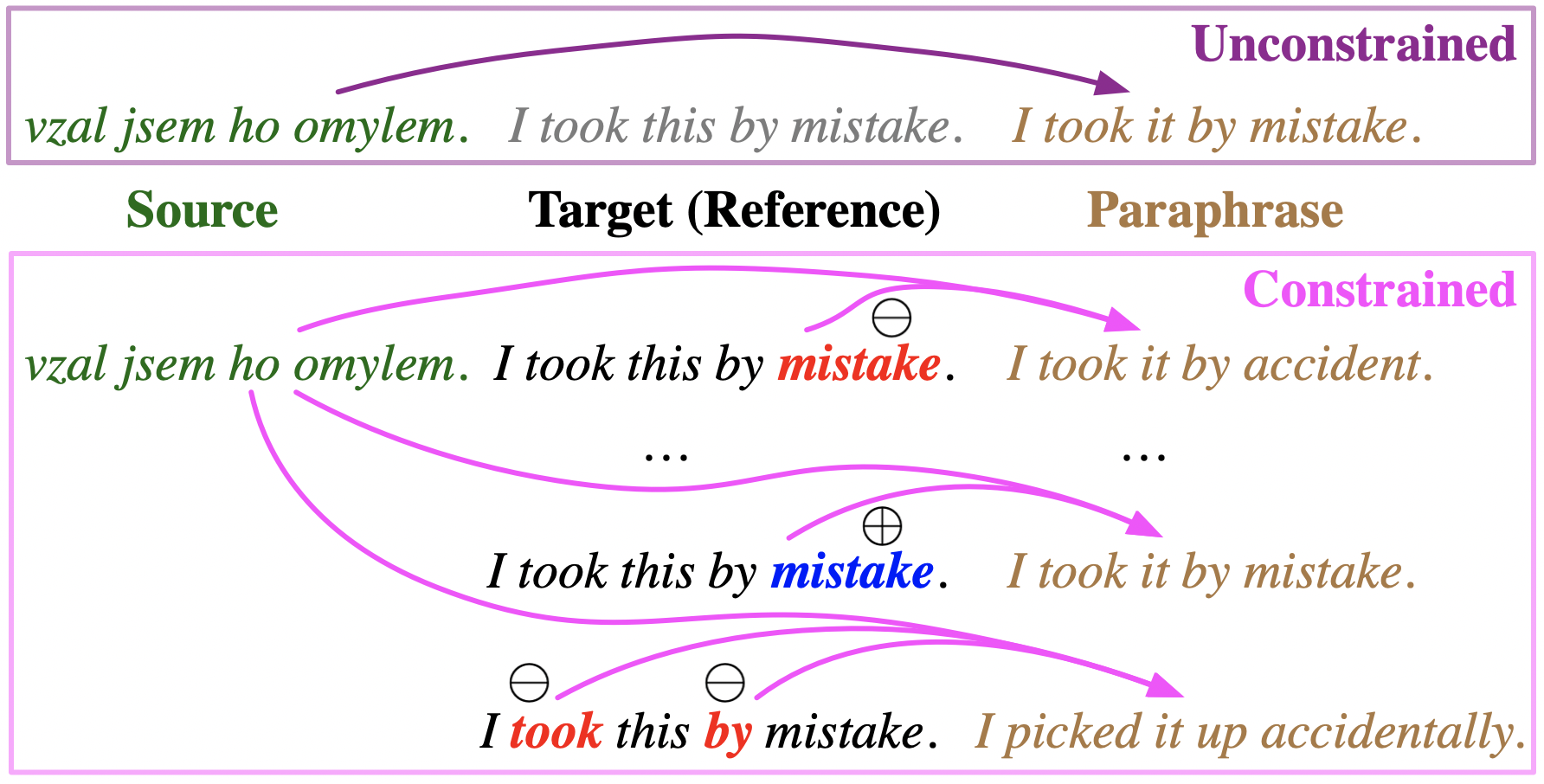}
  \caption{Contrasting prior work (e.g., \paranmt) on building sentential paraphrase collections via translation, {\sc ParaBank} conditions on both the source and \emph{target} side of a translation pair, employing positive and negative lexical constraints derived from the reference to result in \emph{multiple}, \emph{diverse} paraphrases.}
  \label{fig:i-took-it}
\end{figure}

Paraphrastic resources exist at many levels of granularity, e.g., WordNet \citep{miller1995wordnet} for word-level and the Paraphrase Database (PPDB) \citep{ganitkevitch2013ppdb} for phrase-level. We are interested in building a large resource for \textit{sentence}-level paraphrases in English.
In this work, we introduce \parabank, the largest publicly-available collection of English paraphrases we are aware of to date.
We follow and extend the approach of \citet{paranmt}, who generate a set of 50 million English paraphrases, under the name {\textsc{ParaNMT-50M}, via neural machine translation (NMT).

A part of \parabank is trained and decoded on the same Czech-English parallel corpus \citep{czeng16:2016} as \paranmt.
However, \parabank not only contains more reference sentences but also more paraphrases per reference than \paranmt with overall improvements in lexical diversity, semantic similarity, and fluency.
We are able to obtain a larger number of paraphrases per sentence by applying explicit lexical constraints to the NMT decoder, requiring specific words to appear (\textit{positive constraint}) or not to appear (\textit{negative constraint}) in the decoded sentence. Using \parabank, we train, and release to the public, a monolingual sentence re-writing system, which may be used to paraphrase unseen English sentences with lexical constraints.

The main contributions of this work are: 
\begin{itemize}
  \item A novel approach to translation-based generation of monolingual bitext;
  \item The largest and highest quality English-to-English bitext resource to date with 79.5 million references and predicted human-judgment scores;
  \item Manual assessment of candidate paraphrases on semantic similarity, across a variety of generation strategies;
  \item A trained and freely released model for English-to-English rewriting, supporting positive and negative lexical constraints.
\end{itemize}

\parabank is available for download at:
\url{http://nlp.jhu.edu/parabank}.

\section{Background}

The following summarizes key background in approaches to monolingual paraphrasing with regard to \parabank, along with the essential prior efforts that enable \parabank to improve on related work.  We first discuss work on sub-sentential resources that may be: hand-curated, automatically expanded from hand-curated, or fully automatically created.  We then describe efforts at gathering or creating monolingual sentential bitexts, or otherwise sentence-to-sentence paraphrastic rewriting.~\nocite{juri} For additional background, we refer readers to \citet{madnani2010generating}.

\subsection{Paraphrasing Resources}

\paragraph{Lexical Resources}

WordNet~\citep{miller1995wordnet} includes manually-curated sub-sentential paraphrases. It groups single words or short-phrases with similar meanings into synonym sets (\textit{synsets}).
Each synset is related to other synsets through semantic relations (e.g., hypernym, entailment)
to allow the construction of hierarchies and entailment relations.

VerbNet~\citep{verbnet} is a manually-constructed lexicon of English verbs. It is augmented with syntactic and semantic usage of its verb sense members. As a paraphrase resource, VerbNet groups verb senses into classes like ``say-37.7-1" or ``run-51.3.2". Members under the same class share a general meaning as noted by the class name.

FrameNet~\citep{framenet} contains
manually annotated sentences classified into
semantic frames. Though designed as a readable reference and as training data for semantic role labeling, FrameNet can be used to construct sub-sentential paraphrases, owing to rich semantic contents like semantic types and frame-to-frame relations.

Many efforts have aimed to automatically expand gold resources, for example: \citet{semantic_taxonomy} augmented WordNet by combining existing semantic taxonomies in WordNet with hypernym predictions and coordinate term classifications;
and ~\citet{framenet+} tripled the lexical coverage of FrameNet by substituting words through PPDB, with verification of quality via crowdsourcing.

\paragraph{Larger Paraphrases}

There is a rich body of work in automatically inducing phrasal, syntactic or otherwise structural paraphrastic resources.
Some examples include: DIRT~\citep{DIRT}, which extracts paraphrastic expressions over paths in dependency trees, based on an extension of the distributional hypothesis, which states that words that occur in the same contexts tend to have similar meanings~\citep{dist_hypo}; \citet{verb_inf_rules} explored learning inference relations between verbs through broader scopes (document or corpus level), resulting in a richer set of cues for verb entailment detections; and PPDB~\citep{ganitkevitch2013ppdb}, a multi-lingual effort to construct paraphrase pairs by linking words or phrases that share the same translation in another language.   Related to the human scoring pursued here for ParaBank evaluation, in PPDB 2.0~\citep{pavlick2015ppdb}, the authors collected annotations via Mechanical Turk to measure the quality of the induced paraphrases, in order to train a model for scoring all the entries in PPDB for semantic adequacy.

\subsection{Monolingual Bitexts}

\paragraph{Manually Created}

Some monolingual bitexts are created for research in text generation~\citep{pang2003syntax,Robin1995generation} where models benefit from exposure to multiple elicitations of the same concept. The utility of these resources is largely limited by their scale, as the cost of creation is high. Other sources include different translations of the same foreign text, which exist for many classic readings. Work has been done on identifying and collecting sub-sentential paraphrases from such sources~\citep{barzilay2001extracting}. However, the artistic nature of literature could result in various interpretations, often rendering this type of resource unreliable.

\paragraph{\paranmt}

\citet{paranmt} leverage the relative abundance of bilingual bitext to generate sentence-level paraphrases through machine translation. This approach trains a neural machine translation (NMT) model from a non-English source language to English over the entire bitext (Czech-English)~\citep{czeng16:2016}, and decodes the source to obtain outputs that are semantically close to the training target. Decoding in \paranmt solely depends on the trained model and the source text with no inputs derived from the target English sentences. The approach in \paranmt exhibits little control over the diversity and adequacy of its paraphrastic sentence pairs: the application of lexical constraints during decoding is a key distinction between \paranmt and the approach described herein. 

\begin{table*}[t!]
\centering
\begin{tabular}{p{2.5cm}p{5.5cm}p{2.3cm}p{5.7cm}}
\toprule
System &    Reference &     Constraints &                                                    Paraphrase \\
\midrule
$\ominus$\snd IDF                           &            How often do earthquakes \redbold{occur}? &                              $\ominus$occur &                          How often are earthquakes happening? \\
$\ominus$\snd \trd IDF                      &  How \redbold{often} do earthquakes \redbold{occur}? &                       $\ominus$occur, often &                         What frequency do earthquakes happen? \\
$\ominus\langle$BOS$\rangle$ \fst token     &            \redbold{How} often do earthquakes occur? &            $\ominus\langle$BOS$\rangle$ How &                          What frequency do earthquakes occur? \\
PPDB equ                                    &            How \redbold{often} do earthquakes occur? &            $\ominus$often $\oplus$frequently &          How \bluebold{frequently} are earthquakes happening? \\
\midrule
PPDB rev                                    &   This \redbold{myth} involves three misconceptions. &              $\ominus$myth $\oplus$mythology &  There are three misconceptions in this \bluebold{mythology}. \\
$\ominus$\fst IDF                           &   This myth involves three \redbold{misconceptions}. &                     $\ominus$misconceptions &                              This myth has three false ideas. \\
$\ominus$\trd IDF                           &   This myth \redbold{involves} three misconceptions. &                           $\ominus$involves &                            The myth has three misconceptions. \\
\midrule
$\ominus$\snd IDF                           &             It didn't mean \redbold{anything}, okay? &                           $\ominus$anything &                                 It didn't mean a thing, okay? \\
$\ominus$\fst IDF                           &             It didn't mean anything, \redbold{okay}? &                               $\ominus$okay &                           It didn't mean anything, all right? \\
$\ominus$\fst, \trd IDF w/ lexical variants &   It didn't \redbold{mean} anything, \redbold{okay}? &  $\ominus$okay,mean, means,meaning, meant &                                    It was nothing, all right? \\
$\ominus$\fst, \snd IDF                     &   It didn't mean \redbold{anything}, \redbold{okay}? &                     $\ominus$okay,anything &                                  It meant nothing, all right? \\
\bottomrule
\end{tabular}
\caption{Examples of different constraint selection methods in \parabank leading to multiple different paraphrases per reference. System names are defined in \tabref{tab:system_description} with full descriptions in \S\ref{sec:selection}. Negative constraints are labeled in red and positive constraints are labeled in blue.}
\label{tab:example}
\end{table*}

\subsection{Lexically Constrained Decoding} \label{sec:lexdec}

Lexically constrained decoding~\citep{hokamp-liu:2017:lexical_constraint} is a modification to beam search for neural machine translation that allows the user to specify tokens and token sequences that must (or must not) appear in the decoder output.
A lexical constraint can be either \emph{positive} or \emph{negative}. 
A positive constraint requires the model to \textit{include} the constrained token or tokens in the output. 
Negative constraints, on the other hand, require the model to \textit{avoid} certain token or tokens in the output. 
The effect of constraints is to cause the system to generate the best decoding (translation or paraphrase) under those constraints. 

Recently, \citet{fast_lexical} proposed a variant of lexically constrained decoding that reduced complexity from linear to constant-time (in the number of constraints).
This allows us to decode hundreds of millions of sentences with constraints in a reasonable amount of time, and forms a key enabling technology for \parabank. 
An implementation of it is included in Sockeye~\citep{sockeye}, which we use for this work.

\subsection{Efficient Annotation of Scalar Labels (EASL)}

\citet{easl} propose an efficient and accurate method of collecting scalar-valued scores from human annotators, called EASL, by combining pairwise ranking aggregation and direct assessment.
In manually evaluating the quality of our system's paraphrases, we adopt an annotator interface based on EASL.
Human annotators are asked to assess paraphrases' semantic similarity to the reference sentence through a combination of direct numerical assessment and pairwise comparison. This mode of evaluation is akin to the method employed by the Workshop on Statistical Machine Translation (WMT) evaluation through an adaptation of TrueSkill\texttrademark{}~\citep{efficient_elicitation}.

\section{Approaches}

\subsection{Training the model}
We use Sockeye~\citep{sockeye} to train the machine translation model, with which we generate paraphrases under different constraint systems. The training data, CzEng 1.7~\citep{czeng16:2016}, is tokenized\footnote{We used spaCy~\citep{spacy2} to tokenize English text, and MorphoDiTa~\citep{morphodita} to tokenize Czech text.}
and processed through Byte Pair Encoding (BPE)~\citep{bpe}.
To reduce the vocabulary size, we tokenized all numbers to digit-level.

The model's encoder and decoder are both 6-layer LSTMs with a hidden size of 1024 and an embedding size of 512. Additionally, the model has one dot-attention layer.
We trained the model on 2 Nvidia GTX 1080Ti for two weeks.

\subsection{Selection of Lexical Constraints}
\label{sec:selection}

Lexical constraints (\S \ref{sec:lexdec}) can directly influence the diversity and sufficiency of the NMT decoder output (i.e., the translation). 
We generate paraphrases of English translations of Czech sentences using different sets of constraints obtained from the English side of the bitext.
These constraints may be positive or negative, and multiple constraints of either type may be combined simultaneously (provided they are consistent).

The tokens on which we base these constraints are the tokens that appear in the reference sentence (or are morphological variants thereof), though in principle any token could be used as a constraint.
To select the constraints from this pool, we experiment with different ways of selecting these constraints from the reference, resulting in 37 experimental system configurations (\tabref{tab:system_description}): one baseline system with no constraints, three with tokens selected positionally, 30 with positive and negative constraints selected via inverse document frequency (IDF), and three additional systems based on PPDB lookups. Here we describe these selection criteria in detail.

\paragraph{IDF Criteria} We compute each token's inverse document frequency (IDF) from the training data. To avoid constraints with misspelled or overly-specialized words, we exclude tokens with an IDF above 17.0 from consideration as lexical constraints.
We also avoid constraints based on the most frequent English words by setting a minimum IDF threshold of 7.0.
These thresholds are heuristics and we leave optimizations to future works. Among the remaining candidates, the constraint token may be selected by the highest IDF, lowest IDF, or randomly.

\paragraph{Prepositions} We make one exception to the minimum IDF threshold in the case of prepositions, which we found fruitful as diversity-promoting constraints (see \figref{fig:i-took-it}). The allowed prepositions are: 
\textit{about}, \textit{as}, \textit{at}, \textit{by}, \textit{for}, \textit{from}, \textit{in}, \textit{into}, \textit{of}, \textit{on}, \textit{onto}, \textit{over}, \textit{to}, and \textit{with}.

\begin{table}[t!]
 \centering
 \small
 \begin{tabular}{|c|c||c|c|}
 \hline
  {\bf Token} & {\bf IDF} & {\bf Token} & {\bf IDF} \\
 \hline
  proud & 11.1 & told & 7.9 \\
  work & 7.4 & them & 6.2 \\
  her & 5.8 & was & 4.3 \\
  for & 3.6 & to & 2.3 \\
  \hline
 \end{tabular}
 \caption{Tokens assessed, along with their IDFs, of the sentence \textit{``I told her I was proud to work for them.''}}
 \label{tab:IDF_example}
\end{table}

\paragraph{Morphological Variants} To discourage trivial paraphrases, some negative constraint systems include morphological variants of the word, and all negative constraint systems exclude capitalization. For positive constraints, we only consider morphological variants for verbs\footnote{We POS-tagged the reference sentence using SpaCy.}, and only one variant of the selected token is used. For all constraints, only lowercased alphabetical tokens are considered.

\paragraph{Positional Constraints} It has been observed that RNN decoders in dialogue systems can be nudged toward producing more diverse outputs by modifying decoding for only the first few tokens \citep{li-EtAl:2016:N16-11}. Motivated by this observation, we include \textbf{positional constraints}, which require that a given constraint apply only at the \textit{beginning} of the sentence (denoted as $\langle$BOS$\rangle$ in Table \ref{tab:system_description}). In particular, we require the first one, two, or three tokens \textit{not} to match the reference translation (i.e., a \textit{negative} constraint).

\paragraph{PPDB Constraints} We also use PPDB 2.0 \citep{pavlick2015ppdb} as a source for introducing \textit{positive} lexical constraints. For each token in the original English sentence that passes the IDF filter (above), we look up its paraphrases in PPDB\footnote{We use the \texttt{ppdb-2.0-lexical-xl} packet downloaded from \url{paraphrase.org}.}. We randomly select up to three lexical paraphrases, one each of the type \texttt{Equivalence}, \texttt{ForwardEntailment}, and \texttt{ReverseEntailment}, if present. We further require the selected lexical paraphrases to coarsely match the original token's POS tag (e.g., any form of verb, etc.) A negative constraint is then added for the original token, and a positive constraint is added for the lexical paraphrase from PPDB. These negative-positive constraint pairs are applied one at a time (i.e., one pair per decoding).

\paragraph{Example of constraint selection} Here we work through the process of selecting lexical constraints to produce a new paraphrase of the sentence \textit{``I told her I was proud to work for them.''}. We follow the rules of system number 18 (as designated in \tabref{tab:system_description}).

First, tokens with only lower-cased alphabetical letters are assessed; they are listed in \tabref{tab:IDF_example} along with their IDF values. After applying the IDF thresholds and exception for prepositions, the following tokens are in the candidate pool, from which we choose tokens to constrain on (listed in descending order of IDF values): \textit{proud}, \textit{told}, \textit{work}, \textit{for}, and \textit{to}.

Under the configuration of \parabank System 18 (\tabref{tab:system_description}), which avoids tokens with the lowest and the second lowest IDF, a negative constraint is generated for the tokens \textit{for}, \textit{For}, \textit{to}, and \textit{To}.
With these constraints applied, the resulting decoded paraphrase is: \textit{``I told her I was really proud of working with them.''}.

More examples are shown in \tabref{tab:example}.

\begin{table}[t!]
 \centering
 \small
 \begin{tabular}{|c|c|c|c|}
 \hline
  No. & $\oplus/\ominus$ &  Token(s) Selected & Lex. \\
 \hline
  1,2,3 & $\ominus$ & \fst, \snd, \trd~ highest IDF\tablefootnote{Highest within the token pool. Same for lowest.} & None \\
  4,5,6 & $\ominus$ & (\fst, \snd), (\snd, \trd), (\fst, \trd) IDF & None \\
  7 & $\ominus$ & (\fst, \snd, \trd) highest IDF & None \\
 \hline
  8,9,10 & $\ominus$ & \fst, \snd, \trd~ highest IDF & All \\
  11,12,13 & $\ominus$ & (\fst, \snd), (\snd, \trd), (\fst, \trd) IDF & All \\
  14 & $\ominus$ & (\fst, \snd, \trd) highest IDF & All \\
 \hline
  15,16,\textbf{17} & $\ominus$ & \fst, \snd, \textbf{\trd~ lowest IDF} & None \\
  18,19,20 & $\ominus$ & (\fst, \snd), (\snd, \trd), (\fst, \trd)  low & None \\
  \textbf{21} & $\ominus$ & \textbf{(\fst, \snd, \trd) lowest IDF} & None \\
 \hline
  22,23,24 & $\ominus$ & 1, 2, 3 random tokens & None \\
  25,26,27 & $\ominus$ & 1, 2, 3 random tokens & All \\
 \hline
  \textbf{28} &  & \textbf{no constraints} &  \\
 \hline
  29,30,\textbf{31} & $\oplus$ & \fst, \snd, \textbf{3rd highest IDF} & Verb\tablefootnote{If the token is a verb, we pick a random lexical variation to include. E.g., we might constrain on one of "taken", "taking", "takes", or "take" if the original token is "took".} \\
 \hline
  32,33,\textbf{34} & $\ominus$ & positional $\langle$BOS$\rangle$: 1, 2, \textbf{3 tokens} & None \\
 \hline
  \textbf{35},\textbf{36},\textbf{37} & $\ominus\oplus$ & \textbf{PPDB equ, fwd, rev entailment} & None \\
  \hline
 \end{tabular}
 \caption{Different system configurations to generate paraphrases. $\oplus$/$\ominus$ designates the type of constraint we impose on the model: \textit{negative} constraints ($\ominus$) are sets of tokens or ngrams which the decoder must \textit{not} include in its output, while \textit{positive} constraints ($\oplus$) are sets of tokens or ngrams \textit{required} in the output. Additional constraints may be included for lexical variations of the selected token(s), as indicated by the Lex. column. Systems in bold fonts are presented here for evaluation. Evaluations of all systems will be made available with the resource.}
 \label{tab:system_description}
 \end{table}

\section{Extension to other bilingual corpora}

Our methods are independent of the source bilingual corpora. We apply the same pipeline to the $10^9$ word French-English parallel corpus (Giga) \citep{WMT2009}, which has different domain coverage than CzEng. This adds an additional 22.4 million English references, resulting in a total of 79.5 million reference sentences for \parabank. We conducted manual evaluation on paraphrases generated from both CzEng and Giga, and included the additional PPDB constraint systems for Giga.
 
\section{Evaluation}

\begin{table*}[t!]
\centering
\begin{tabular}{ccl|lllll|lllll}
\toprule
&     \# &     System  & \multicolumn{5}{c}{Czech-English} & \multicolumn{5}{c}{French-English} \\
&         &             &  len=5 &  len=10 &  len=20 &  len=40 &  Avg. &len=5 &  len=10 &  len=20 &  len=40 &  Avg. \\
\midrule
\parbox[t]{2mm}{\multirow{9}{*}{\rotatebox[origin=c]{90}{Semantic Similarity}}} &   - &         \textsc{ParaNMT} &   73.47      & 73.49      &    75.49      & 71.09      & 73.39       &-&-&-&-&- \\
&   17 &  $\ominus$\trd low IDF &   71.59      & \bf{75.71} &    \bf{79.67} & \bf{77.31} & \bf{76.07} &73.64 &               72.69 &               83.60 &               80.15 &    77.52  \\
&   21 &    $\ominus$3 low IDF &   59.86      & 66.56      &    70.76      & 69.77      & 66.74      &62.84 &               69.56 &               81.73 &               77.08 &    72.80  \\
&   28 &         no con. &   \bf{76.63} & \bf{78.35}* &    \bf{83.19}* & \bf{80.35} & \bf{79.63}* &79.09 &               74.28 &               84.86 &               83.02* &    80.31*  \\
&   31 &  $\oplus$\trd top IDF &   \bf{77.12}* & \bf{75.22} &    \bf{82.98} & \bf{79.91} & \bf{78.81} &80.22* &               73.78 &               85.01 &               80.02 &    79.76  \\
&   34 &  $\ominus$first 3 tks &   \bf{74.03} & \bf{77.74} &    \bf{81.88} & \bf{81.04}* & \bf{78.67} &78.12 &               74.37* &               84.14 &               82.74 &    79.84 \\
\cmidrule(lr{1em}){2-13}
&35 &        PPDB Equ &  -&-&-&-&-&           79.22 &               67.30 &               82.09 &               79.98 &    77.15 \\
&  36 &        PPDB Fwd &   -&-&-&-&-&           65.40 &               69.39 &               85.10* &               82.75 &    75.66 \\
&  37 &        PPDB Rev &   -&-&-&-&-&           64.25 &               66.28 &               76.35 &               72.49 &    69.84 \\
  \midrule\midrule
\parbox[t]{2mm}{\multirow{9}{*}{\rotatebox[origin=c]{90}{Lexical Diversity}}}& 35 & \textsc{ParaNMT} &     18.32 &   25.49 &   32.25 &   33.84 &    27.48 &        - &         - &         - &         - &          - \\
& 17 &  $\ominus$3rd low IDF &  \bf{9.25} &\bf{20.62}&   35.76 &   41.88 &\bf{26.88}&     6.59 &     15.57 &     26.78 &     34.54 &      20.87 \\
& 21 &    $\ominus$3 low IDF & \bf{0.00*}  &\bf{7.84*}&\bf{22.04*}&\bf{28.90*}&\bf{14.70*}&     1.21* &      6.90* &     18.28 &     22.02 &      12.10 \\
& 28 &         no con. &  19.32     &   28.41 &   42.02 &   46.63 &    34.10 &    14.07 &     21.12 &     30.57 &     38.66 &      26.11 \\
& 31 &  $\oplus$3rd top IDF &  19.09     &   29.16 &   41.42 &   46.96 &    34.16 &    15.28 &     21.69 &     31.49 &     37.56 &      26.51 \\
& 34 &  $\ominus$first 3 tks & \bf{13.22} &\bf{24.90}&   39.11 &   44.67 &    30.47 &    10.87 &     18.15 &     27.70 &     36.87 &      23.40 \\
\cmidrule(lr{1em}){2-13}
& 35 &        PPDB equ &      - &       - &       - &       - &        - &     4.46 &     13.33 &     25.88 &     29.54 &      18.30 \\
& 36 &        PPDB fwd &      - &       - &       - &       - &        - &     1.51 &      8.87 &     14.14* &     19.03* &      10.89* \\
& 37 &        PPDB bkw &      - &       - &       - &       - &        - &     3.94 &      9.69 &     18.94 &     26.24 &      14.70 \\
\bottomrule
\end{tabular}
\caption{
\textbf{Top}: Semantic similarity between paraphrases and reference sentences, as scored by human annotators on a 0-100 scale (least to most similar). Results are grouped by length of reference sentences \{5, 10, 20, 40\}.
System names and numbers correspond to \tabref{tab:system_description}.
Improvements over \textsc{ParaNMT} (Czech-English only) in bold. Asterisk (*) indicates best in column.
\textbf{Bottom}: Lexical diversity between generated paraphrases and reference sentences, as computed by a modified BLEU score with no length penalty. Results are grouped by length of reference sentences, and BLEU is computed over concatenated references and concatenated paraphrases. Lower BLEU scores indicate greater lexical divergence; the lowest per column (bottom half) is indicated by (*).
}
\label{tab:mega}
\end{table*}

We evaluate the quality of paraphrases by both semantic similarity and lexical diversity. A good paraphrase should strive to preserve as much meaning as possible while using lexically diverse expression. Otherwise, the result is either a trivial re-write or fails to convey the original meaning. We understand these two metrics as interdependent and sometimes conflicting -- a high lexical diversity likely sacrifices semantic similarity. The goal of \parabank is to offer not only a balance between the two, but also options across the spectrum for different applications. Of course, good paraphrases should also be fluent in their expression, so we also evaluate paraphrases for grammaticality and meaningfulness, independent of their reference.

For brevity, we picked 5 \parabank systems (bold in \tabref{tab:system_description}) to cover negative, positive, positional, and no constraint. We also include system 21 (3 lowest IDF tokens) to show that too many constraints might significantly hurt semantic similarity.
Full evaluations on all proposed systems will be included with the release of the resource.

\subsection{Scoring \parabank paraphrases}

Following the approach of PPDB 2.0 \citep{pavlick2015ppdb}, we trained a supervised model on the human annotations we collected. We extracted several features from reference-paraphrase pairs to predict human judgments of semantic similarity on all paraphrases, with the exception of those whose reference contains more than 100 tokens post-BPE. The regression model achieves reasonable correlation with human judgment on the test data with a Spearman's $\rho$ of 0.53 on CzEng and 0.63 on Giga.

\subsection{Baseline comparison} \label{sec:baseline}

Our baseline system with no lexical constraints applied shows substantial improvement compared to \paranmt. This could be a combination of improved training data and NMT framework. \paranmt is trained on a 51.4M subset of CzEng1.6, while \parabank used CzEng1.7, a 57.0M subset of CzEng1.6. We also switched to \textsc{Sockeye} as our training framework. 
This baseline improvement gives us more flexibility to pursue explicit lexical diversity with reasonable compromise in semantic similarity.

\subsection{Semantic similarity} \label{sec:mturk}

Human judgment remains the gold standard of semantic similarity. We randomly sampled 100 Czech-English sentence pairs from each of the four English token lengths: 5, 10, 20, and 40. We translate 400 sentences from CzEng under 34 \parabank systems (without PPDB constraints) and 400 sentences from Giga under 37 \parabank systems. Then, we merge identical outputs and add in the corresponding \paranmt entries to the CzEng evaluation.

We randomize the paraphrase pool and formulate them into Mechanical Turk Human Intelligence Tasks. Inspired by the interface of EASL \citep{easl}, we ask workers to assign each paraphrase a score between 0 and 100 by adjusting a slider bar. Each worker is presented with one reference sentence and five attempted paraphrases at the same time. Occasionally, we present workers the reference sentence itself as a candidate paraphrase and expect it to receive a perfect score. Workers who fail to do so more than 10\% of all times are disqualified for inattentiveness. 
In total, we incorporated the annotations of 44 workers who contributed at least 25 judgments.
Each paraphrase receives independent judgments from at least 3 different workers.

We then calculate the average score for each sentence pair, before averaging over all pairs from each \parabank system (or \paranmt). The final score for each system is a number between 0 and 100. 

The top half of \tabref{tab:mega} shows the average human judgment over 100 sentences per reference length for \parabank systems and \paranmt, grouped by sentence length.

Best performing \parabank systems from each reference length outperform \paranmt relatively by 5.0\%, 6.6\%, 10.2\%, and 14.0\% in terms of semantic similarity (corresponding to 5, 10, 20, 40 tokens per reference sentence).

\subsection{Lexical diversity}

We used a modified BLEU score to evaluate lexical diversity and \textbf{a lower score suggests a higher lexical diversity}. Specifically, we concatenate\footnote{After switching all tokens to lowercase and stripping punctuation to avoid rewarding trivial re-writes.}
multiple paraphrastic sentence pairs into one reference paragraph and one paraphrase paragraph, and calculate the associated BLEU score \textbf{without brevity penalty}. This modification ensures that we don't reward shorter paraphrases. We compared the result with a naive unigram precision metric and they show a strong correlation with a Spearman's $\rho$ of 0.98.

The bottom half of \tabref{tab:mega} shows this modified BLEU score for each \parabank system and \paranmt, grouped by reference length. For every length, there is at least one \parabank system that exhibits higher lexical diversity than \paranmt; unsurprisingly, the \parabank systems that apply the greatest number of lexical constraints tend to yield the greatest lexical diversity (e.g., system 21).

\subsection{Meaningfulness and grammaticality}

We ask annotators to comment on each paraphrase's fluency by flagging sentences that are completely nonsensical or indisputably ungrammatical. We consider a sentence nonsensical or ungrammatical when at least one independent annotator flags it as so.

We then calculate the percentage of sentences that are deemed both meaningful and grammatical for each \parabank system and \paranmt.

The result is shown in \tabref{tab:fluency}. System 34 (avoid first 3 tokens) shows a 12.6\% improvement over \paranmt. In all, 21 out of 34 proposed \parabank systems contain a smaller proportion of nonsensical or ungrammatical sentences than \paranmt. The full set of annotations are available with the resource.

\begin{table}[tbp]
\centering
\begin{tabular}{llll}
\toprule
\# &     System & Cz-En & Fr-En \\
\midrule
 - &\textsc{ParaNMT} &         73.25 &              - \\
17 &  $\ominus$\trd low IDF &    \bf{76.50} &          73.75 \\
21 &    $\ominus$3 low IDF &         65.50 &          65.00 \\
28 &         no con. &    \bf{81.75} &         80.00* \\
31 &  $\oplus$\trd top IDF &    \bf{74.00} &          72.75 \\
34 &  $\ominus$first 3 tks &   \bf{82.50}* &          77.50 \\
35 &        PPDB equ &             - &          69.97 \\
36 &        PPDB fwd &             - &          71.19 \\
37 &        PPDB rev &             - &          52.46 \\
\bottomrule
\end{tabular}
\caption{Percentage of paraphrases for each system that are rated by human annotators as both grammatical \textit{and} meaningful, independent of similarity to the reference sentence. Improvements over \textsc{ParaNMT} (Czech-English only) in bold. Asterisk (*) indicates best in column.
}
\label{tab:fluency}
\end{table}

\section{Constrained monolingual rewriting}

\cite{joshua_rewriter} explored sentential rewriting with machine translation models. Inspired by their work, we use a subset of \parabank, with more than 50 million English paraphrastic sentence pairs (English text from CzEng as source, \parabank outputs as target), to train a monolingual NMT model, and decode with the same types of constraint systems.
We present the following result as a proof of concept that highlights the potential for and problems with the most straightforward instantiation of the model. A thorough investigation of building such a monolingual model is outside the scope of this work.

We decide to use the same LSTM model instead of more advanced self-attention models to contrast between the bilingual and monolingual models. After training for one epoch, we decode the model with the same 5 constraint systems (no. 17, 21, 28, 31, 34) evaluated for the bilingual model, and ask human annotators\footnote{Same setup as \S \ref{sec:mturk}. The result includes 8 workers who contributed more than 25 judgments.} to compare their semantic similarity to the reference sentence in the same way. We sampled 100 sentences across the same 4 lengths (25 sentences per length); each sentence receives at least 3 independent judgments. The semantic similarity scores for this monolingual system are reported in \tabref{tab:mono} (``Monolingual'') alongside lexical diversity scores (modified BLEU).

\begin{table*}[tbp]
\centering
\begin{tabular}{llll|ll}
\toprule
\# &          System & \multicolumn{2}{c}{Semantic Similarity} & \multicolumn{2}{c}{Lexical Diversity} \\
    &                 &           \multicolumn{1}{c}{Bilingual (Cz-En)} &  \multicolumn{1}{c}{Monolingual} &         \multicolumn{1}{c}{Bilingual (Cz-En)} &  \multicolumn{1}{c}{Monolingual} \\
\midrule
  - &\textsc{ParaNMT} &               73.89 (std. 4.60) &               -          &        27.36 (4.89)  &      -                    \\
 17 &  $\ominus$3rd low IDF &               76.79 (std. 3.77) &        74.56 (std. 9.17) &             27.20 (std. 14.67) &        33.04 (std. 18.58) \\
 21 &    $\ominus$3 low IDF &               66.48 (std. 8.25) &        63.35 (std. 19.59)&             14.28 (std. 13.35) &        24.82 (std. 20.52) \\
 28 &         no con. &               81.91 (std. 3.11) &        86.25 (std. 3.94) &             33.90 (std. 11.58) &        44.36 (std. 13.69) \\
 31 &  $\oplus$3rd top IDF &               80.64 (std. 3.60) &        83.09 (std. 4.23) &             33.48 (std. 12.66) &        43.80 (std. 13.39) \\
 34 &  $\ominus$first 3 tks &               80.04 (std. 1.95) &        84.47 (std. 4.10) &             30.20 (std. 13.50) &        39.81 (std. 10.52) \\
  - &       Reference &               99.81 (std. 0.16) &        -         &            100.00 (std. 0.00)  &       -          \\
\bottomrule
\end{tabular}
\caption{
Comparison of bilingual (Czech-English) and monolingual (English-English) paraphrasing systems in terms of (1) \textbf{semantic similarity} as rated by human annotators on a scale of 0-100, and (2) \textbf{lexical diversity} as measured by a modified BLEU score without length penalty, where lower BLEU scores are taken as evidence of greater lexical diversity. We observe that similarity and diversity scores for the monolingual rewriting systems exhibit higher variance than bilingual systems (sample standard deviations given in parentheses); however, the monolingual rewriter is able to generate English paraphrases in the absence of a Czech reference sentence.
}
\label{tab:mono}
\end{table*}

Outputs from the monolingual model show a significant boost in semantic similarity compared to bilingual counterparts, system 28 (no constraint) shows an improvement of 16.7\%. This is accompanied by an increase in BLEU score, a sign of less lexical diversity.  Example outputs from the monolingual model can be found in \tabref{tab:mono_examples}.

\begin{table}[t!]
 \centering
 \small
 \begin{tabular}{|c|}
 \hline
  {\bf Reference} \\

  \textit{Hey, it's nothing to be ashamed of.} \\
 \hline
 \hline
  {\bf Paraphrases from the monolingual model} \\
  \textit{Hey, it's nothing to be embarrassed.} \\
  \textit{Hey, it's nothing to be ashamed.} \\
  \textit{Hey, it's not like you're ashamed of.} \\
  \textit{Hey, you don't have to worry about that.} \\
  \textit{Hey, you don't have to be ashamed.} \\
  \textit{Hey, there's nothing you can be ashamed of.} \\
  \textit{You don't have to be ashamed of it.} \\
  \textit{Hey, there's nothing you can do about it.} \\
  \textit{Oh, hey, it's no big deal.} \\
  \hline
 \end{tabular}
 \caption{Example paraphrases generated from the monolingual rewriting model, after applying the same set of lexical constraints described in \tabref{tab:system_description} and merging duplicates.}
 \label{tab:mono_examples}
 \end{table}

As evidenced in our examples, some monolingual systems may generate slightly more nonsensical or ungrammatical sentences than their bilingual counterparts: future work will pursue more extensive model training and data filtering for the monolingual model.  Our intent here is to foremost illustrate the quality of {\sc ParaBank} as a resource, while illustrating the feasibility of training and employing a monolingual sentence rewriting model built atop the {\sc ParaBank} artifact.

\section{Conclusions and Future Work}
We created \parabank by decoding a neural machine translation (NMT) model with lexical constraints. We applied our methods to CzEng 1.7 and Giga, leading to a large collection of paraphrases with 79.5 million references and on average 4 paraphrases per reference, which we make available for download at \url{http://nlp.jhu.edu/parabank}. Via large-scale crowdsourced annotations, we found the overall best performing \parabank system exhibits an 8.5\% relative improvement in terms of semantic similarity over prior work. Analysis on lexical diversity showed the potential of \parabank as a more diverse and less noisy paraphrastic resource. In addition to releasing hundreds of millions of English sentential paraphrases, we also release a free, pre-trained, model for monolingual sentential rewriting, as trained on {\sc ParaBank}.

With the existence of {\sc ParaBank} and an initial monolingual rewriting model, future work can investigate how more advanced NMT models, such as those with self-attention structures, can lead to better rewriting. One may also investigate the automatic expansion of resources for a variety of NLP tasks. For example, in Machine Translation one might create sentential paraphrases from the English side of bitexts for low-resource languages: cases where only small numbers of gold translations exist in English, and are expensive or otherwise problematic to expand by hand.  In Information Extraction, one may rewrite sentences with structured annotations such as for Named Entity Recognition (NER), with  positive constraints that phrases representing known NER spans be preserved while some tokens of the remainder be negatively constrained, thereby providing additional novel sentential contexts for IE system training. In educational NLP technology, one might wish to rewrite a sentence that includes or excludes target vocabulary words a language learner does not understand or is trying to acquire. There are many other such examples in NLP where the ability to rewrite existing datasets with lexical constraints could lead to significantly larger and more diverse training sets, with no additional human labor.
To pursue such work may require a large, high quality monolingual bitext to train a rewriting model, and an NMT decoder supporting both positive and negative constraints, such as we have introduced here.

\section*{Acknowledgments}
We thank Chu-Cheng Lin and three anonymous reviewers for their thoughtful feedback. This research was supported in part by DARPA AIDA. The views and conclusions contained in this publication are those of the authors and should not be interpreted as representing official policies or endorsements of DARPA or the U.S. Government.

\bibliography{ref}
\bibliographystyle{aaai}
\end{document}